# Design and Experimental Validation of a 3D-Printed Torsional Series Elastic Actuator for Safe Human–Robot Interaction

**Joel Hidalgo Pisco[1], Melissa Cobos Condo[1], Luigi Miranda[1], Dennys Paillacho[1]**
[1] ESPOL Polytechnic University
Guayaquil, Ecuador
joezhida@espol.edu.ec; mncobos@espol.edu.ec; luidamir@espol.edu.ec; dpaillac@espol.edu.ec

***Abstract -*** Ensuring intrinsic safety in physical human-robot interaction (pHRI) is a critical requirement for social and service robots. While Series Elastic Actuators (SEAs) offer hardware-based compliance, traditional metallic designs often require complex, multi-part assemblies. This paper presents the design, finite element analysis (FEA), and experimental validation of a low-stiffness, torsional spring for SEAs, manufactured via 3D-printed thermoplastic polyurethane (TPU). The compliant element exhibits a highly linear torque-deformation response (Ks = 0.066 Nm/degree), matching numerical predictions with under 3% deviation, a variance attributed to FDM structural anisotropy. To accommodate external interactions using standard position-limited servomotors, a hybrid position controller with torque-threshold switching was implemented. Experimental evaluations demonstrate the system's ability to accurately track non-stationary trajectories and safely yield to external disturbances. Furthermore, the inherent material damping of the TPU acts as a passive low-pass filter, preventing high-frequency oscillations during control mode transitions. The proposed architecture offers a cost-effective, reliable, and easily manufacturable solution for safe pHRI.

## 1 Introduction

As robots increasingly perform everyday activities in shared environments such as care assistance [1], rehabilitation [2], and education [3] ensuring intrinsic safety during physical human-robot interaction (pHRI) has become a critical priority [4]. While software-based control strategies, including collision detection and impedance control [5], [6], enable robots to react to unexpected impacts, they often lack the inherent physical adaptability of hardware-based solutions. Consequently, Series Elastic Actuators (SEAs) have emerged as a prominent approach to achieving intrinsic safety [7]. By introducing a compliant element between the motor and the load, SEAs decouple the actuator's inertia, reduce the apparent stiffness of the joint, and passively absorb impacts, making them highly suitable for collaborative workspaces. This work focuses on the development and implementation of a low-stiffness SEA intended for the upper-limb joints of a social robotic platform. In service robots, arms constitute the primary mechanism for gestural interaction; joints such as the elbow are fundamental for enabling natural movements during gestures like waving or handshaking. Integrating a compliant element into these joints reduces the apparent inertia and mitigates the risks of abrupt impacts, ensuring safe and user-friendly physical interaction. Furthermore, the stiffness requirements for this application differ significantly from those commonly adopted in industrial manipulators or exoskeleton systems, where highly rigid elastic elements are typically employed to withstand elevated loads [8], [9]. In contrast, the proposed design intentionally incorporates a highly compliant profile. This behavior is justified by the nature of the intended application, where human safety is prioritized over high force transmission capabilities, as well as by the mechanical limitations of the selected Dynamixel MX-28 actuator. Under these conditions, maximizing the compliance of the robotic arm to absorb external disturbances becomes more valuable than maintaining structural rigidity, allowing the arm to yield smoothly if the user interrupts or opposes the motion.

## 2 Related Works

### 2.1 SEA configurations

SEA can be constructed using a variety of materials, shapes, mechanical components, and configurations. Among these configurations, there are specific designs that utilize linear tension springs, as well as others that incorporate torsional springs. The latter are tailored in their design based on the intended application and are widely used due to their monolithic nature, unlike systems with linear springs, which require multiple components.

Several rotary SEAs utilize configurations based on tension linear springs to absorb tangential forces and achieve bidirectional compliance. For instance, arrangements uniformly distributed along a circular perimeter using six linear springs [10], [11] or four linear springs [12] have been implemented to provide symmetric stiffness responses through antagonistic tension-compression behavior. While effective, these linear spring networks typically require multiple moving components, which inherently increases mechanical complexity, packaging constraints, and system assembly space.

To overcome the spatial and component limitations of linear systems, monolithic torsional springs have been widely adopted. These single component architectures facilitate direct interaction torque estimation solely through the angular deformation of the elastic element. Various geometric configurations have been explored to optimize compliance and energy absorption, including corrugated coil units [13], optimized disc geometries [8], and sinusoidal wave-shaped architectures modeled via analytical beam theory [14]. Additionally, non-planar configurations, such as cylindrical compliant elements incorporating repeated U-shaped circumferential patterns, have been proposed to optimize the balance between torsional compliance and compact sizing during physical interaction [15]. However, these metallic geometries often rely on highly specialized, expensive manufacturing processes to prevent severe stress concentrations under peak loads.

Beyond conventional architectures, advanced compliant mechanisms have been developed to achieve tailored or variable stiffness profiles. These include mechanisms combining S-shaped cantilever beams with specialized cam profiles for high torque resolution under low loading conditions [16], and multi-part systems utilizing rotating camshafts interacting with flexible outer ring teeth to localize bending deformation [9]. Furthermore, integrated optimization frameworks have been established to balance structural stiffness with fatigue life and thermal performance in long term robotic operations [17]. While these specialized mechanisms successfully adapt the actuator's torque-deformation behavior, their multipart nature introduces mechanical backlash and assembly friction, highlighting the need for simplified, high compliance monolithic solutions manufactured via cost-effective methods.

### 2.2 Materials and procedures of manufacturing

Similar to geometric configurations, the materials and manufacturing processes for SEAs exhibit significant variability based on required stiffness, energy absorption, fatigue resistance, and geometric constraints. As summarized in Table 1, rigid metallic alloys (e.g., AISI 4041 or 18Ni maraging steel) are predominantly used and manufactured via subtractive methods like waterjet cutting or wire EDM [8], [13], [18], [19]. While these processes ensure high dimensional accuracy and fatigue resistance, they involve complex fabrication. In contrast, polymer-based materials such as CNC-machined TPU enable larger elastic deformations and provide a more compliant, damped mechanical response, making them highly attractive for safe human–robot interaction (HRI) [20].

Table 1: Materials and manufacturing process used for SEA fabrication

| Spring | Material | Manufacture Process |
|---|---|---|
| [8] | AISI 4041 | Water-jet |
| [13] | Martensite Steel 300 | Wire |
| [20] | Polyurethane Thermoplastic (TPU) | CNC |
| [19] | AISI 18 Ni | - |

To achieve this compliant behavior, this work presents the design, finite element analysis (FEA), and experimental validation of a torsional spring for SEAs, utilizing TPU manufactured via 3D printing. The proposed actuator is designed for robotic arm applications involving HRI, where compliant behavior and safe physical interaction are essential. The methodology combines FEA and experimental characterization to evaluate the stiffness and mechanical response of the compliant element, leveraging TPU due to its flexibility, energy absorption capability, and suitability for compliant actuation systems.

## 3 Methodology

This section describes the design procedure of the proposed actuator, including the determination of the target stiffness, the numerical analysis of the compliant element, and the experimental characterization of the fabricated spring.

The design process of the compliant element was based on the methodology proposed by [8] and [21], with several modifications introduced to adapt the procedure to the requirements of the present application. Initially, the target spring stiffness was determined according to the maximum torque that can be delivered by the selected servomotor.

The procedure used to define the geometry of the compliant element consisted of identifying a configuration capable of satisfying the design requirements listed in Table 2. Subsequently, the proposed geometries were optimized and evaluated through numerical simulations in order to identify the configuration exhibiting the most suitable mechanical response in terms of stiffness, deformation, and stress distribution.

Finally, the selected compliant element was fabricated using 3D printing and experimentally characterized to compare the measured stiffness with the values predicted through numerical simulations. The overall workflow followed during the design process is presented in Fig. 1.

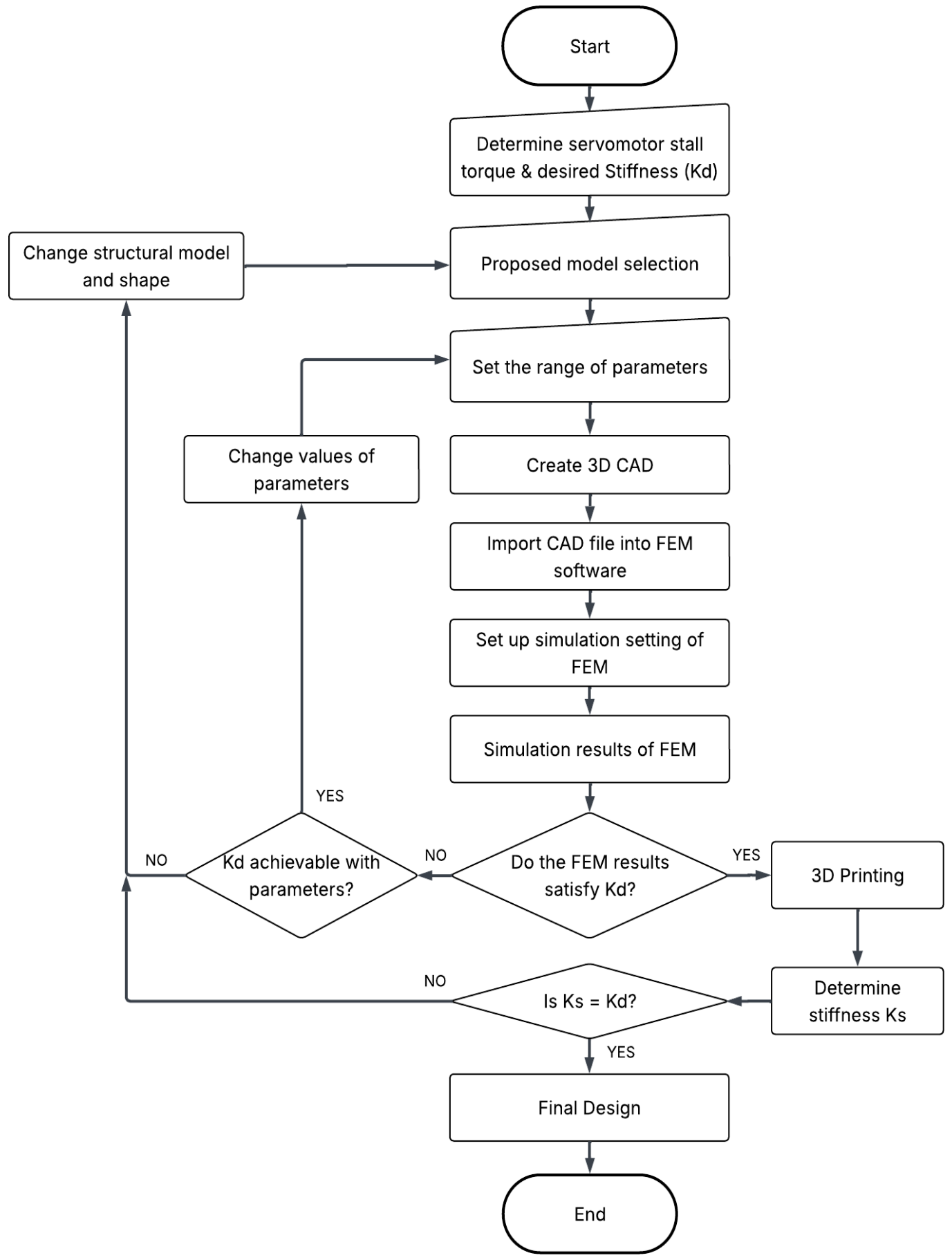

Fig. 1: Flowchart of the torsional spring design

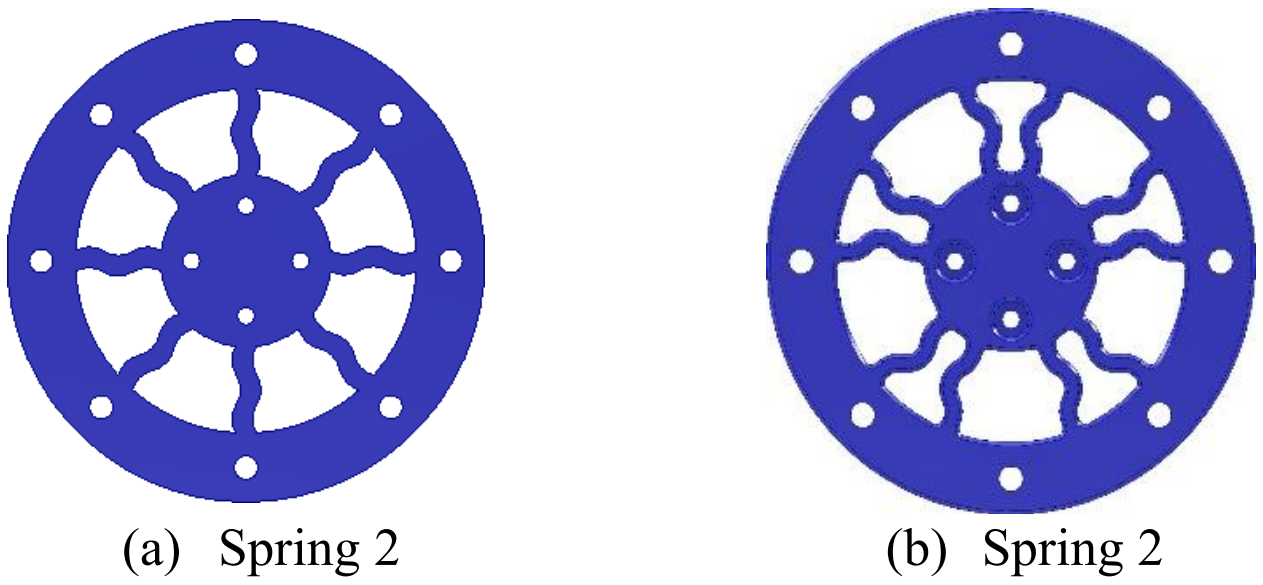
(a) Spring 2 (b) Spring 2
Fig. 2: Proposed torsional compliant mechanisms for SEA

### 3.1 Stiffness Requirement Definition

The design process begins with the definition of the initial system requirements, specifically establishing the maximum torque of the selected servomotor. This torque was determined experimentally by subjecting the actuator to a variable torsional load. The loads were progressively applied using masses of 50 g positioned at a distance of 0.05 m from the rotation axis, thereby producing the corresponding torque on the actuator shaft. The maximum admissible load was identified when the servomotor was no longer capable of sustaining the applied torque and yielded under the imposed force.

To ensure conservative operating conditions, the experimentally obtained maximum torque was reduced by applying a safety factor of 0.7, thereby preventing the servomotor from operating at its maximum load capacity. In addition, a maximum angular deformation constraint of 10◦ was imposed when the flexible element is subjected to the defined maximum torque. By applying Hooke's law for rotational systems, defined as $T = Kd*\theta s$, the variable Kd representing the desired theoretical stiffness of the mechanism, was determined. This parameter establishes the stiffness required for the actuator to satisfy the prescribed safety and motion transmission criteria.

$$K_d = 0.7 * \frac{T_{max}}{10}\left[\frac{\mathrm{N}}{\mathrm{degree}}\right] \quad (1)$$

### 3.2 Torsional Spring Parametrization and Design

The proposed spring geometry must provide a predictable, linear torque-deformation relationship and bidirectional symmetric compliance to ensure stable control during interactive tasks. To achieve the target stiffness defined in Table 2, two spring geometries (Fig. 2) were modeled and evaluated through FEA. The specific geometric parameters varied during this optimization process are detailed in Table 3.

The numerical evaluation focused on angular deformation, equivalent Von Mises stress distribution, and structural boundaries under peak loads. During evaluation, Model 1 presented severe stress concentrations at the roots of the coils connected to the central ring, as well as excessively large deformations under maximum load, compromising its structural integrity. In contrast, Model 2 exhibited equivalent stresses well within the material's elastic range and its deflection remained strictly within the prescribed operational limits. Consequently, Model 2 demonstrated superior structural performance and mechanical reliability under bidirectional torsional loading.

Table 2: Design parameters of the compliant element

| Design parameter | Desired parameter value |
|---|---|
| Maximum output torque | 0.64 Nm |
| Desired stiffness | 0.064 Nm/degree |
| External diameter | 70 mm |
| Inner diameter | 25 mm |
| Maximum deflection | 10 degrees |
| Spring material | TPU |

Table 3: Range of variable parameters

| | Parameter | Minimum value | Maximum value | Step |
|---|---|---|---|---|
| Spring 1, 2 | Number of spires | 6 | 10 | 1 |
| | coil thickness | 2 mm | 4 mm | 0.4 mm |

### 3.3 Finite Element Analysis

FEA was conducted using Ansys to evaluate the mechanical behavior of the proposed geometries. The compliant element was discretized using tetrahedral solid elements, culminating in a simulation model of approximately 80,260 elements and 127,030 nodes after a mesh convergence analysis. Based on manufacturer specifications, the TPU material was modeled with a Young's modulus of 26 MPa, a Poisson's ratio of 0.48, and a density of 1220 Kg/m3.

Boundary conditions replicated the physical setup: one end was fully constrained while a maximum operational torsional load was applied to the opposite end. Torsional stiffness (Ks) was estimated directly from the resulting angular deformation (θs) using Hooke's law (Ks = T/θs).

### 3.4 Manufacturing of the Compliant Element

TPU was selected as the manufacturing material due to its high elasticity, exceptional passive impact absorption capabilities, and excellent manufacturability through fused deposition modeling (FDM). Unlike rigid metallic alloys commonly used in industrial SEAs, the inherent viscoelasticity and damping properties of TPU are highly advantageous for a robotic arm designed for social interactions, such as human greeting gestures, where disturbance accommodation and collision mitigation are strict design priorities.

Finally, once the stiffness predicted through the finite element method (FEM) showed satisfactory agreement with the desired stiffness value, the proposed spring was fabricated using additive manufacturing techniques. The compliant element was manufactured through 3D printing using an Ender 3 V2 printer equipped with a 0.4 mm nozzle. The printing parameters employed during the fabrication process are summarized in Table 4.

Table 4: Parameters of 3D printing

| Parameter | Dimension |
|---|---|
| Layer height | 0.2 mm |
| Wall line | 3 |
| Infill Pattern | Grid |
| Infill Density | 50% |
| Printing Temperature | 230 °C |
| Build Plate Temperature | 25 °C |

### 3.5 Control Architecture

A hybrid position controller with torque-threshold switching was developed to track desired trajectories while preventing mechanical overload (Fig. 4). Furthermore, due to the presence of the flexible element, the real position of the arm is equal to the sum of the angular position of the motor plus spring deformation:

$$\theta_{arm} = \theta_m + \theta_s \quad (2)$$

Under nominal conditions ($\tau \leq \tau0$), a standard PID controller regulates the angular position based on the estimated tracking error (Fig. 3a). However, because the servomotors used in the system operate strictly in position control modes, active impedance control cannot be directly implemented.

To achieve adaptive compliance, if the estimated interaction torque exceeds the safety threshold ($\tau0$), the system temporarily suspends strict reference tracking and activates a torque compensation mode (Fig. 3b). In this mode, the motor's reference position is dynamically adjusted in the direction of the spring's deformation to absorb the external impact and decrease the transmitted effort. This switching logic is defined as:

$$u_{control} = \begin{cases} u_{PID} & \text{If } \tau \leq \tau_0 \\ u_{comp} & \text{If } \tau > \tau_0 \end{cases} \quad (3)$$

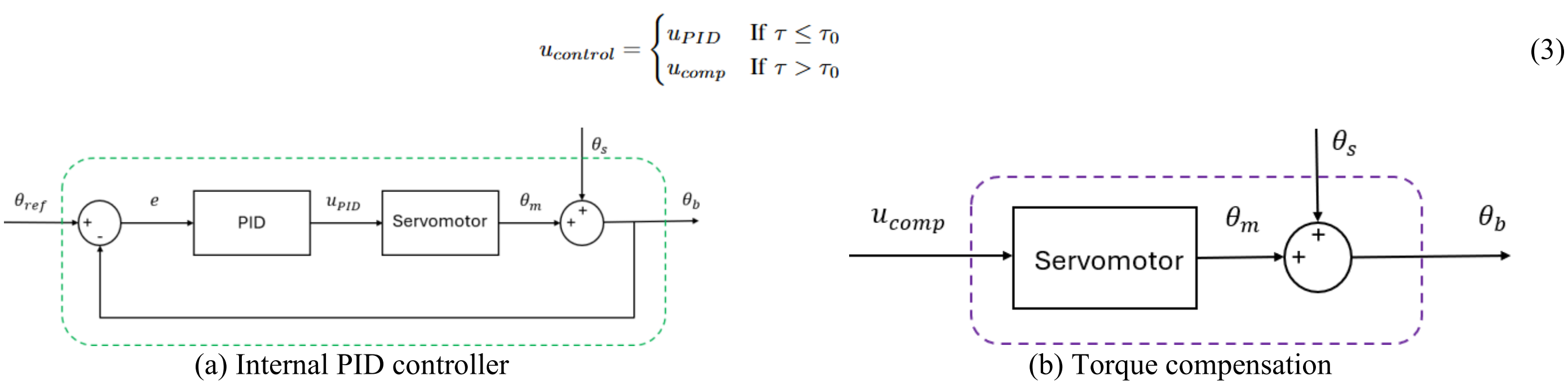

(a) Internal PID controller (b) Torque compensation

Fig. 3: Control system block diagrams.

## 4 Results

### 4.1 Spring Stiffness

To determine the experimental stiffness of the torsional spring. Known loads were progressively applied to a 0.05 m radius flywheel to generate torque and induce deformation. Specifically, the applied mass was increased in increments of 50 g, and for each load step, the corresponding angular deflection was measured using an encoder with a resolution of 0.3 degrees. Based on this experimental data, a torque-deformation curve was constructed, and a linear curve fitting procedure was performed.

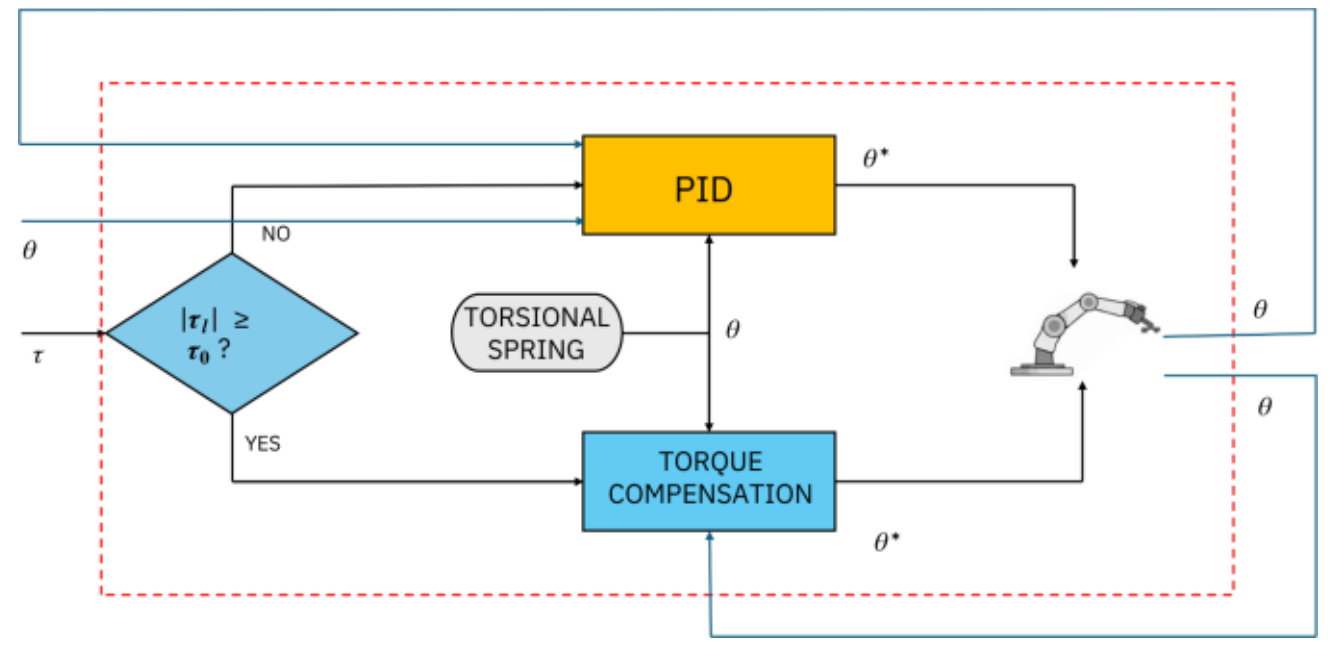

Fig. 4: Hybrid position controller with torque threshold switching

The torsional stiffness was determined as the slope of this function via least-squares regression. The experimental trials yielded a physical stiffness of Ks = 0.066 Nm/degree. Correspondingly, the FEA predicted a numerical stiffness of Ksim = 0.0656 Nm/degree. Both values were evaluated against the initial theoretical target of Kd = 0.064 Nm/degree established during the design parametrization. As illustrated in Fig. 5, both the simulated and experimental torque-deformation curves exhibit a highly proportional and linear response (R2 > 0.99) throughout the evaluated 10◦ operational range.

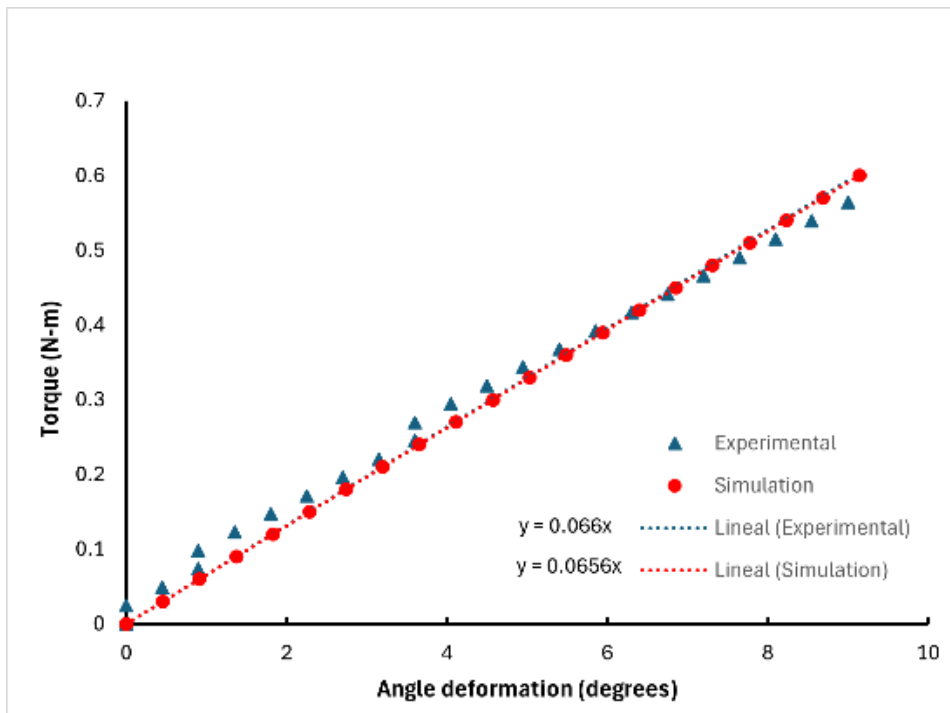

Fig. 5 Experimental and simulated torque-deformation characteristics of the compliant element. The highly proportional response (R2>0.99) validates the predictability of the TPU spring for interaction torque estimation

## 4.2 Experimental Evaluation of the Control System

*1)* System response following a fixed reference: In this stage, the proposed controller was evaluated on a one-degree-of-freedom robotic arm by imposing a constant angular reference, as shown in Fig. 6a. In the absence of external disturbances, the system converges to the reference position with a stable response, consistent with the previously characterized behavior of the internal PID controller.

When an external disturbance is introduced, with a magnitude that generates a torque in the elastic element exceeding the predefined threshold Fig. 6b, the system exhibits a modified dynamic response. In particular, the arm yields under the applied force, resulting in a displacement from the nominal reference. This behavior is consistent with the implemented control strategy, in which the corrective action is attenuated or modified when the torque measured in the series elastic actuator exceeds the established limit. Once the disturbance is removed and the torque returns within the allowable range, the controller restores its nominal action, driving the system back to the reference position.

Furthermore, as observed in Figs 6a and 6b, trajectory deviation occurs only when the torque estimated from the deformation of the elastic element exceeds the thresholds defined in the control law. This confirms that the change in system dynamics is not due to controller instability, but rather to the implemented protection and adaptation logic. Such a strategy enables the system to exhibit compliant behavior during external interactions, while preserving its ability to recover the desired reference position.

*2)* System response following a variable reference: For this evaluation, a sinusoidal angular reference was employed to analyze the dynamic performance of the controller under non-stationary trajectories. Fig. 7 presents the system response when the servomotor operates at its maximum speed in the absence of external disturbances. It can be observed that the arm trajectory

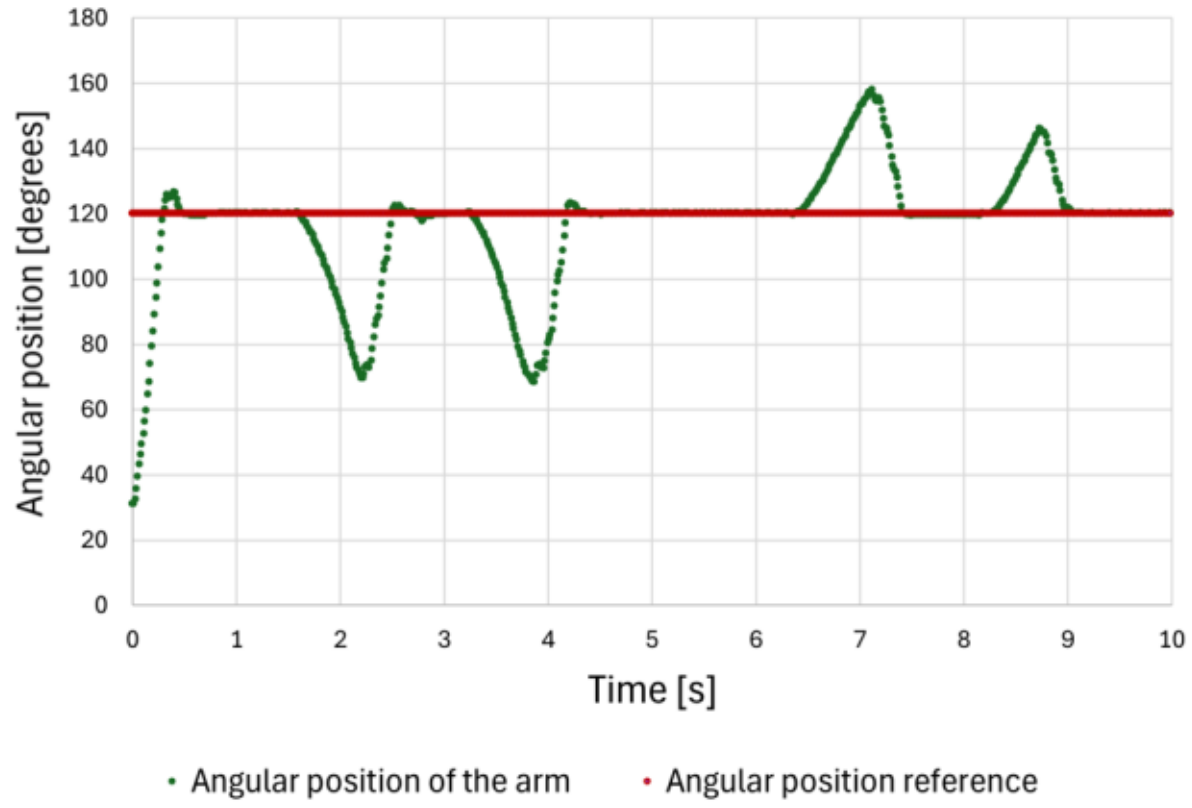

(a) Dynamic response of the robotic arm tracking a constant angular reference. The system successfully yields and deviates from the setpoint only when the external disturbance exceeds the predefined safety boundaries.

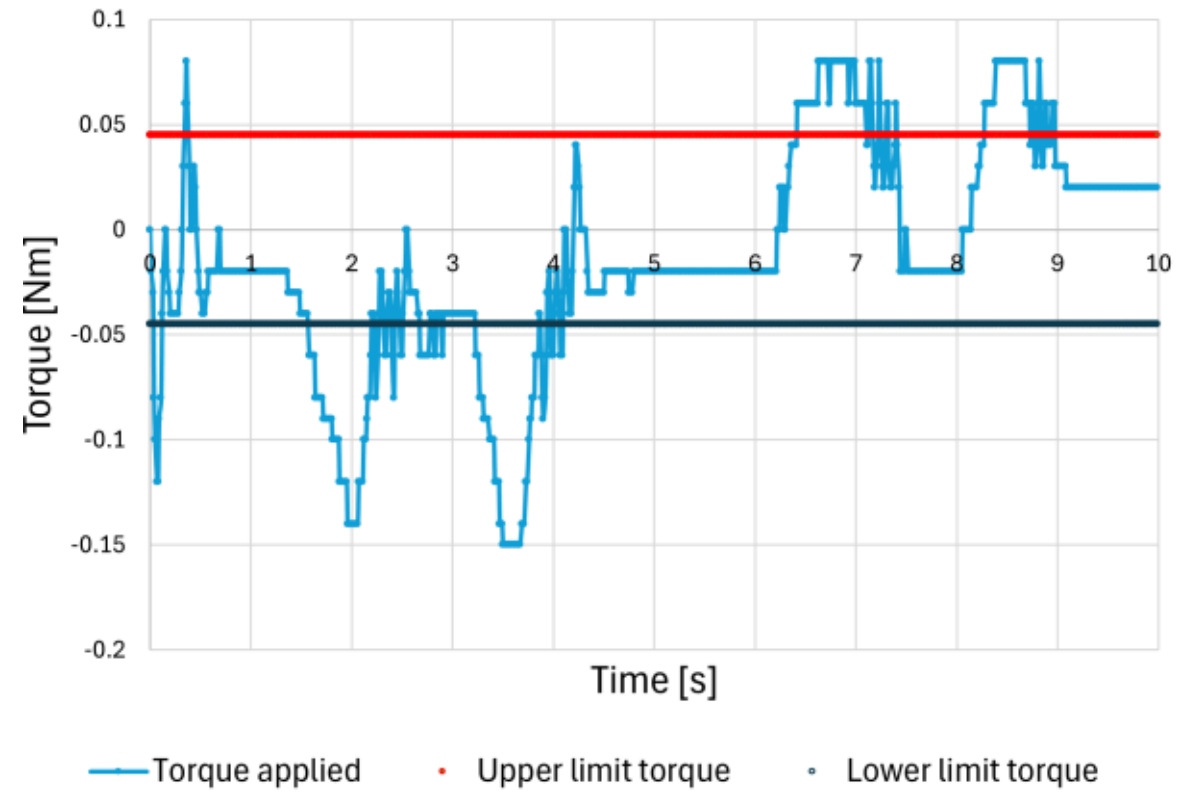

(b) Temporal evolution of the interaction torque during the fixed reference task, demonstrating the activation of the compensation mode when the applied torque surpasses the upper and lower limits.

Fig. 6: System performance under a fixed reference with external forces, showing the dynamic response (a) and the corresponding interaction torque

(Green curve) closely follows the sinusoidal reference (blue curve), exhibiting a low tracking error and minimal phase lag. This behavior demonstrates that, under nominal conditions and with high actuation speed available, the controller is capable of accurately reproducing time varying reference signals.

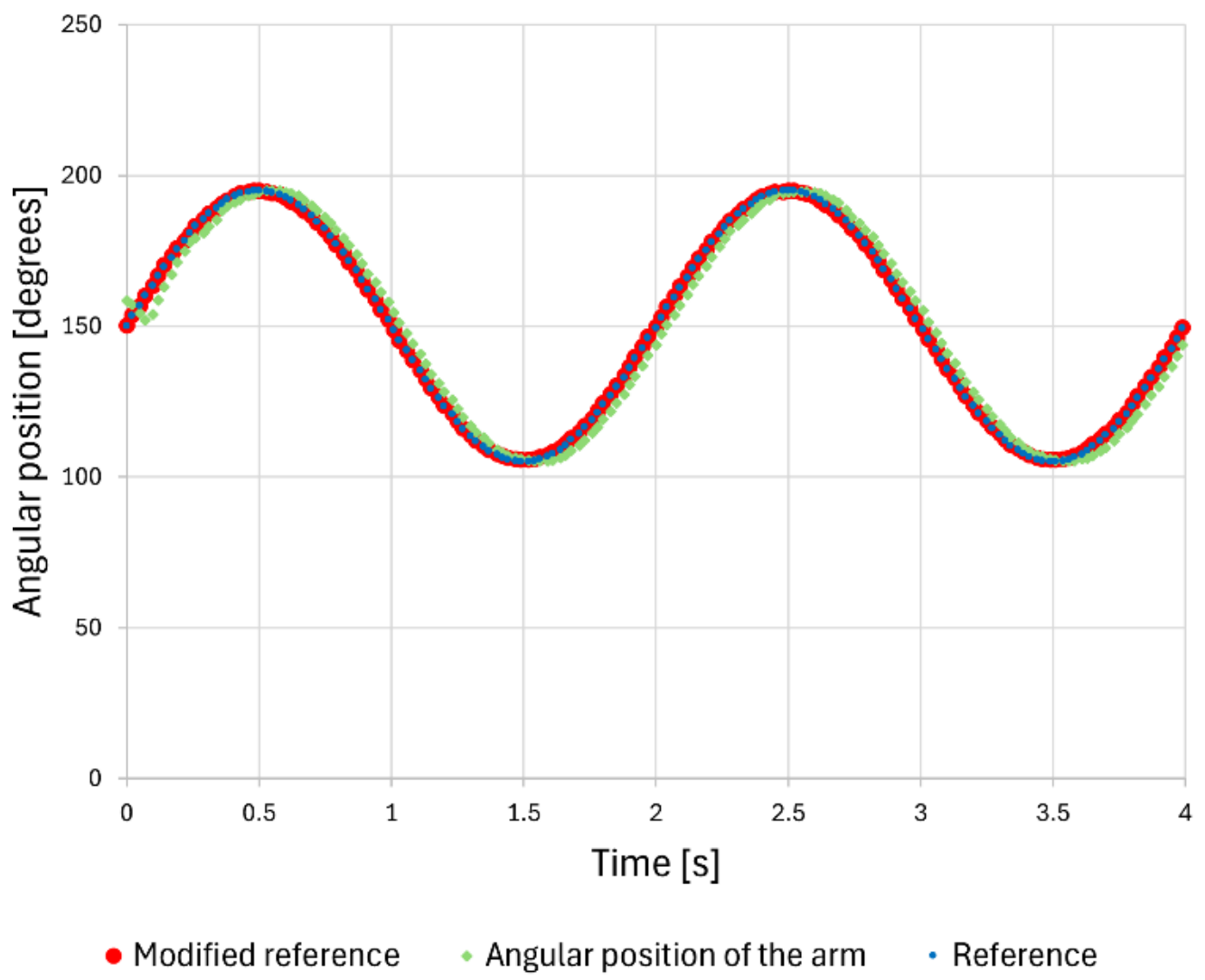

Fig. 7: Trajectory tracking performance of the hybrid controller under nominal conditions without external disturbances. The system exhibits accurate tracking of the sinusoidal reference with minimal phase lag.

Fig. 8a shows the system's behavior at maximum speed in the presence of an external disturbance. When the torque measured in the elastic element exceeds the upper threshold, the reference signal is modified by the control strategy (orange curve), resulting in a temporary trajectory adaptation aimed at limiting mechanical interaction. It is important to note that the modified reference respects the imposed physical limit of 210°, preventing the system from exceeding its safe operating range.

Despite the disturbance induced alteration, the actuator maintains adequate tracking of the modified reference. Once the disturbance vanishes and the torque returns to the allowable range, the modified reference gradually converges back to the original sinusoidal signal.

Fig. 8b complements this analysis by showing the temporal evolution of the torque, where the threshold exceedance is clearly observed, triggering the modification of the servomotor trajectory.

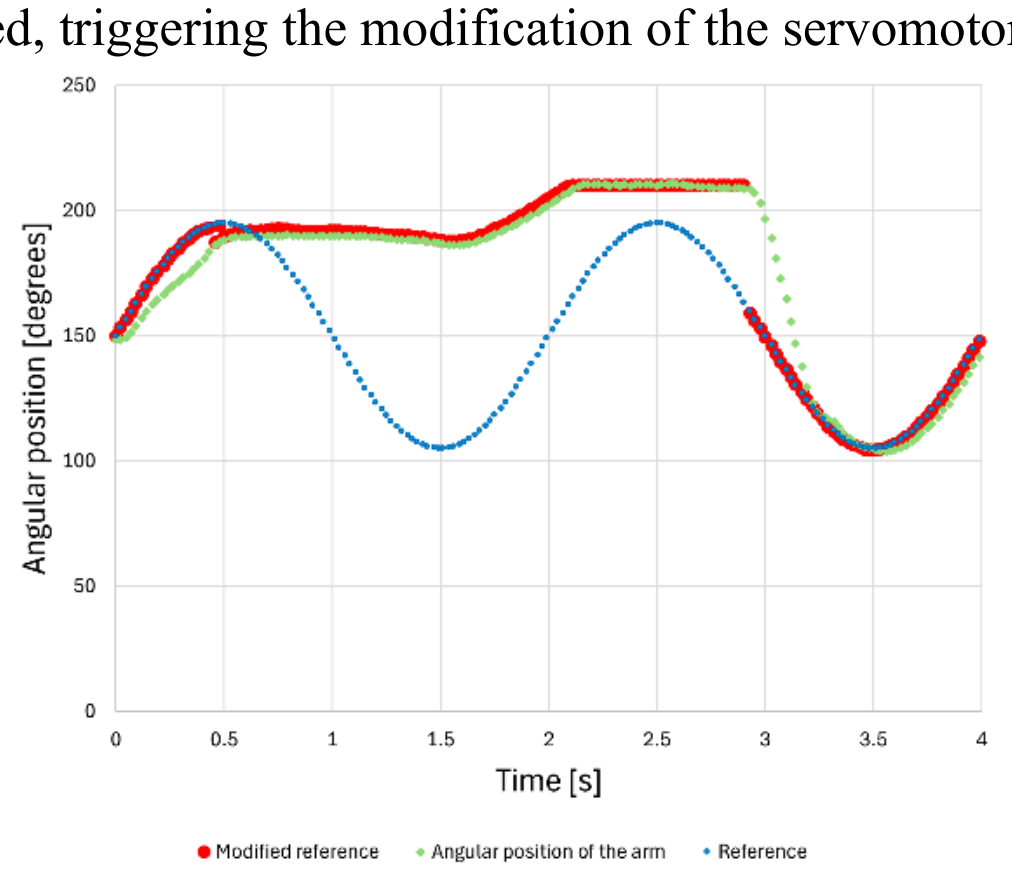

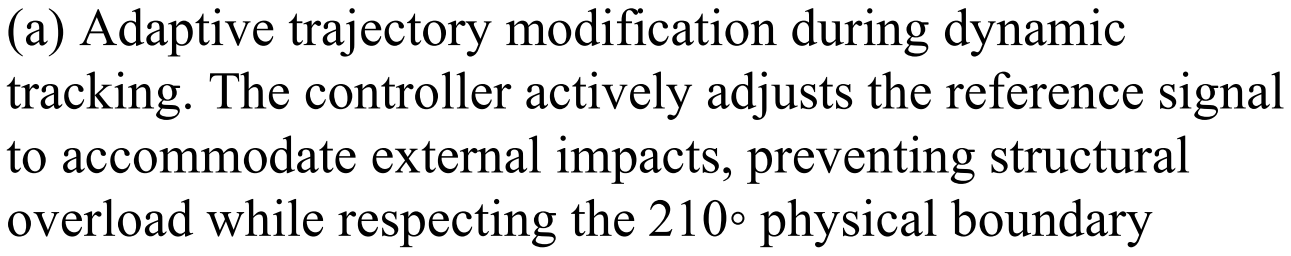

(a) Adaptive trajectory modification during dynamic tracking. The controller actively adjusts the reference signal to accommodate external impacts, preventing structural overload while respecting the 210◦ physical boundary

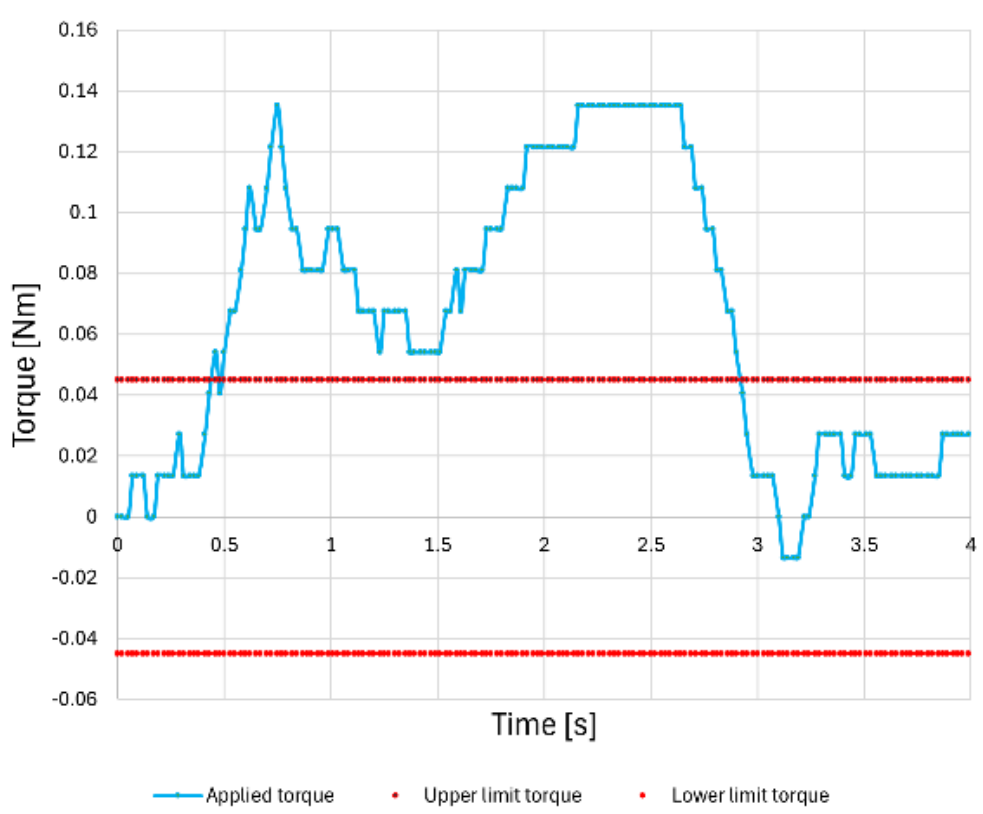

(b) Torque profile during the variable reference task, illustrating the precise moments the threshold switching logic is triggered to prioritize collision mitigation.

Fig. 8: System performance under a variable reference with external forces, detailing the adaptive trajectory modification (a) and the corresponding torque profile (b).

## 5 Conclusion

This work presented the design, finite element analysis, and experimental validation of a monolithic torsional compliant element for SEAs. By leveraging FDM and TPU, the proposed methodology provides a cost-effective and highly adaptable mechanical solution specifically tailored for low stiffness robotic applications. The main conclusions of this study are: The manufactured compliant element exhibits a linear torque-deformation response (Ks = 0.066 Nm/degree), demonstrating that 3D-printed TPU can provide reliable compliance without the need for complex, multi-part metallic assemblies. Minor discrepancies (less than 3%) between the numerical model and the physical prototype were successfully linked to the structural anisotropy inherent to the FDM process. The implementation of a hybrid position controller with torque-threshold switching proved highly effective for disturbance accommodation. The system maintained accurate trajectory tracking under nominal conditions and successfully yielded to external forces when torque boundaries were exceeded. The natural viscoelasticity and damping properties of the TPU material acted as a passive low-pass filter. This prevented high frequency oscillations during control mode commutations, ensuring stable and smooth transitions. Future work will focus on integrating advanced hyperelastic material models into numerical simulations to fully capture the viscoelastic effects of TPU and scaling the proposed SEA architecture to multi degree of freedom manipulators for complex physical interaction tasks.